\documentclass[a4paper,12pt]{article}

\usepackage{natbib}
\usepackage[ansinew]{inputenc}
\usepackage{amsmath}    
\usepackage{amssymb}
\usepackage{graphicx}  

\begin{document}

\def\argmax{\mathop{\rm argmax}}

\title{\LARGE Kernel Regression by Mode Calculation of the Conditional Probability Distribution}
\date{}

\author{\small Steffen Kühn\\ \small Technische Universit\"at Berlin\\ \small Chair of Electronic Measurement and Diagnostic Technology\\ \small Einsteinufer 17, 10585 Berlin, Germany\\ \small steffen.kuehn@tu-berlin.de}

\maketitle

\begin{abstract}
The most direct way to express arbitrary dependencies in datasets is to estimate the joint distribution and to apply afterwards the argmax-function to obtain the mode of the corresponding conditional distribution. This method is in practice difficult, because it requires a global optimization of a complicated function, the joint distribution by fixed input variables. This article proposes a method for finding global maxima if the joint distribution is modeled by a kernel density estimation. Some experiments show advantages and shortcomings of the resulting regression method in comparison to the standard Nadaraya-Watson regression technique, which approximates the optimum by the expectation value.
\end{abstract}

\section{Introduction}

Regression is a very important method in engineering and science for the estimation of the dependencies between two or more variables on the basis of some given sample points. The best known regression method is certainly the parametric regression technique after Legendre and Gauss, which minimizes the squared error between a model -- often a polynom -- and the samples. 

The least squares method is fast and well suitable for strongly linearly correlated data, but seldom appropriate for high-dimensional problems with difficult, unknown, and non-linear dependencies. For these problems, non-parametric regression techniques -- like kernel or Nadaraya-Watson regression methods -- are more suitable (\cite{Nadaraya64,Watson64}). The first step of Nadaraya-Watson regression is to estimate the unknown joint density distribution $p_{\boldsymbol{X},\boldsymbol{Y}}(\boldsymbol{x},\boldsymbol{y})$ of the given sample data $D = \{(\boldsymbol{x}_1,\boldsymbol{y}_1),\ldots,(\boldsymbol{x}_n,\boldsymbol{y}_n)\}$ by a kernel-density estimator \citep{Scott92}. The resulting model distribution has in the most general case the form 
\begin{equation}
\tilde{p}_{\boldsymbol{X},\boldsymbol{Y}}(\boldsymbol{x},\boldsymbol{y}) = \sum\limits_{i=1}^{m} a_i\,\phi_i(\boldsymbol{x} - \boldsymbol{x}_i)\psi_i(\boldsymbol{y} - \boldsymbol{y}_i)
\label{f0}
\end{equation}
with $m \leq n$ and $\sum_{i=1}^{m} a_i = 1$. Furthermore, the kernel functions $\phi_i$ and $\psi_i$ have to be normalized so that the integrals over all values for $\boldsymbol{x}$ and $\boldsymbol{y}$ are one. With this model, the conditional distribution $\tilde{p}_{\boldsymbol{Y},\boldsymbol{X}}(\boldsymbol{y}|\boldsymbol{x})$ can be easily derived:
\begin{equation}
\begin{split}
\tilde{p}_{\boldsymbol{Y}|\boldsymbol{X}}(\boldsymbol{y}|\boldsymbol{x}) &= \frac{\tilde{p}_{\boldsymbol{X},\boldsymbol{Y}}(\boldsymbol{x},\boldsymbol{y})}{\tilde{p}_{\boldsymbol{X}}(\boldsymbol{x})} = \frac{\tilde{p}_{\boldsymbol{X},\boldsymbol{Y}}(\boldsymbol{x},\boldsymbol{y})}{\int\limits_{\boldsymbol{y}} \tilde{p}_{\boldsymbol{X},\boldsymbol{Y}}(\boldsymbol{x},\boldsymbol{y}) \mathrm{d}\boldsymbol{y}} \\
& = \frac{\sum\limits_{i=1}^{m} a_i\,\phi_i(\boldsymbol{x} - \boldsymbol{x}_i)\psi_i(\boldsymbol{y} - \boldsymbol{y}_i)}{\sum\limits_{i=1}^{m} a_i\,\phi_i(\boldsymbol{x} - \boldsymbol{x}_i)}.
\label{f1}
\end{split}
\end{equation}
This distribution represents the relative probabilities for realizations of $\boldsymbol{y}$ given a vector $\boldsymbol{x}$. But for a regression, we do not need a probability distribution, but a single vector. The most intuitive choice is, of course, the mode of the conditional distribution, that means the value $\boldsymbol{y}$ for which $\tilde{p}_{\boldsymbol{Y}|\boldsymbol{X}}$ becomes maximal. For this case, the regression function $\tilde{f}(\boldsymbol{x})$ takes the specific form 
\begin{equation}
\tilde{\boldsymbol{y}} = \tilde{f}(\boldsymbol{x}) = \argmax\limits_{ \boldsymbol{y}}\{\tilde{p}_{\boldsymbol{Y}|\boldsymbol{X}}(\boldsymbol{y}|\boldsymbol{x})\}.\label{f2}
\end{equation}
The difficulty is, however, that the maximization is not easy to calculate, because the expression~(\ref{f1}) is highly non-linear. 

On the other hand, the expected value of $\tilde{p}_{\boldsymbol{Y}|\boldsymbol{X}}(\boldsymbol{y}|\boldsymbol{x})$ regarding $\boldsymbol{y}$ is easy to calculate:
\begin{equation}
\int\limits_{\boldsymbol{y}} \boldsymbol{y}\,\tilde{p}_{\boldsymbol{Y}|\boldsymbol{X}}(\boldsymbol{y}|\boldsymbol{x})\, \mathrm{d}\boldsymbol{y}= \frac{\sum\limits_{i=1}^{m} a_i\,\boldsymbol{y}_i\,\phi_i(\boldsymbol{x} - \boldsymbol{x}_i)}{\sum\limits_{i=1}^{m} a_i\,\phi_i(\boldsymbol{x} - \boldsymbol{x}_i)}.
\end{equation}
The idea of Nadaraya and Watson was to approximate expression~(\ref{f2}) by
\begin{equation}
\tilde{\boldsymbol{y}} \approx \frac{\sum\limits_{i=1}^{m} a_i\,\boldsymbol{y}_i\,\phi_i(\boldsymbol{x} - \boldsymbol{x}_i)}{\sum\limits_{i=1}^{m} a_i\,\phi_i(\boldsymbol{x} - \boldsymbol{x}_i)},
\end{equation}
which is often sufficiently good. But there are also some potential problems.

\begin{figure}[th]
\centerline{\includegraphics[width=\columnwidth]{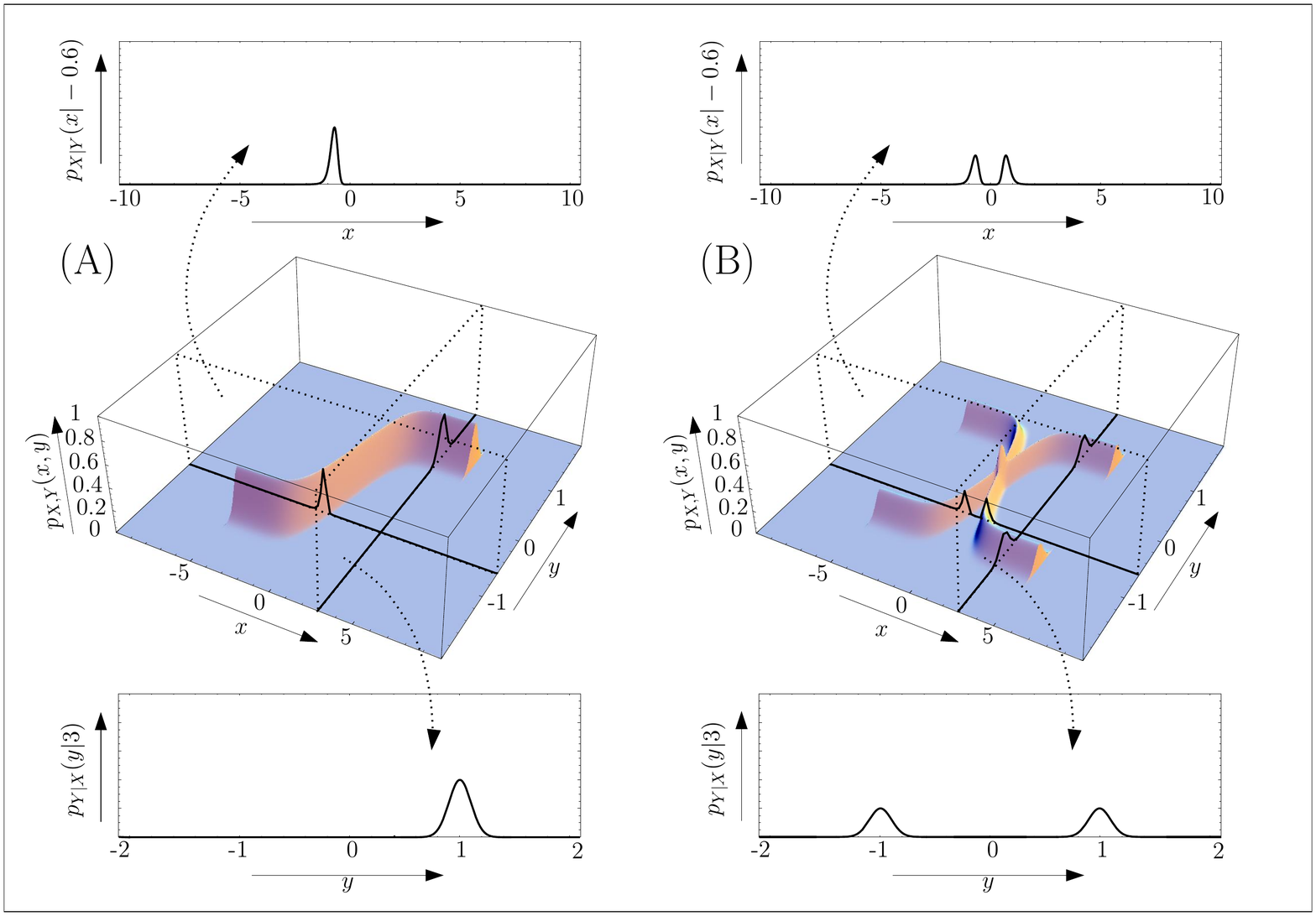}}
\caption{Joint distributions with unambiguous and ambiguous relations between $x$ and $y$.}
\label{fig1}
\end{figure} 

Figure~\ref{fig1} demonstrates this for two different joint distributions. Case (A) is uncritical and describes essentially a hyperbolic tangent function. Case (B), however, causes problems. Here, the variables $x$ and $y$ are non-linearly correlated too, but the underlying dependency cannot be described by a function. The difficulty becomes obvious when considering the conditional distributions $p_{X|Y}(x|-0.6)$ and $p_{Y|X}(y|3)$, which are no longer unimodal. A computation of the expected value would yield in both cases zero, which is far away from probable values for $y = -0.6$ or $x = 3$.

In contrast to the Nadaraya-Watson method, the calculation of expression~(\ref{f2}) should not lead to problems. The $\argmax$-calculation cannot resolve the ambiguousness, of course, due to the fact that two global maxima exist, but it is better to return only \textit{one} maximum than a completely incorrect value. Especially for high-dimensional tasks, this shortcoming of the Nadaraya-Watson method can be annoying, because the occurrence of ambiguousness is difficult to recognize. To overcome this ``curse of compromise'', the next section proposes a method for solving expression~(\ref{f2}) directly if the density estimation is given in the form~(\ref{f0}).

\section{Finding Global Maxima for Kernel Density Estimations}

The fundamental idea of the here proposed method to find the global maximum of a probability density function is to utilize its special properties. In general, the \textit{global} maximum of an arbitrary function, which is, for example, given as a piece of code, can be found only by trial and error. In principle, this can be also applied to find the global maximum of a probability density function. But this ``blind'' search would be very inefficient and the remaining likelihood to find still better values does not become zero, regardless of how long the algorithm runs.

But for probability densities $h(\boldsymbol{y})$, this remaining likelihood can be reduced very fast by using $h$-distributed sample points, instead of evenly distributed samples. Why is it so? To answer this question, we assume that $q$ accordingly distributed sample points $\boldsymbol{y}_i$ with $i=1,\ldots,q$ have been generated. For each $\boldsymbol{y}_i$, there is a percentage $\alpha_i$ for more improbable realizations $\boldsymbol{y}$. Let $\alpha_i$ be $99\%$. The probability that all other $q-1$ generated samples have lower $\alpha$-values is $(99\%/100\%)^{q-1}$. For $q = 10000$, this probability is only $2.27\,10^{-44}$! In practice, this means that it is impossible not to come close to the global maximum with $10000$ $h$-distributed sample points. That is all.

Fortunately, the generation of accordingly distributed samples is not very difficult for kernel density estimations like expression~(\ref{f0}). In the first step, we insert for the calculation of expression~(\ref{f2})  the given value $\boldsymbol{x}$ and get 
\begin{equation}
\tilde{\boldsymbol{y}} = \argmax\limits_{\boldsymbol{y}}\left(h(\boldsymbol{y})\right)
\label{f4}
\end{equation}
with
\begin{equation}
h(\boldsymbol{y}) := \sum\limits_{i=1}^{m} b_i\,\psi_i(\boldsymbol{y} - \boldsymbol{y}_i)
\label{f6}
\end{equation}
and
\begin{equation}
b_i := \frac{a_i\,\phi_i(\boldsymbol{x} - \boldsymbol{x}_i)}{\sum_{j=1}^{m} \phi_j(\boldsymbol{x} - \boldsymbol{x}_j)}.
\end{equation}
Note that $h$ fulfills the requirements for a probability density function. Furthermore, the $b_i$ can be interpreted as probabilities\footnote{Many $b_i$ are very low for a given value $\boldsymbol{x}$. The corresponding kernels should be omitted to improve the computation speed.} because of $\sum_{j=1}^{m} b_j = 1$. 

In the next step, we generate a dataset 
\begin{equation}
D' = \{\boldsymbol{y}_1',\ldots,\boldsymbol{y}_q'\} \label{f5}
\end{equation}
of $h$-distributed random samples. For this purpose, we can utilize the distribution function $H$ of the density function $h$:
\begin{equation}
\begin{split}
H(\boldsymbol{y}) = \int\limits_{-\boldsymbol{\infty}}^{\boldsymbol{y}} \sum\limits_{i=1}^{m} b_i\,\psi_i(\boldsymbol{z} - \boldsymbol{y}_i) \mathrm{d}\boldsymbol{z} & =  \sum\limits_{i=1}^{m} b_i\,\int\limits_{-\boldsymbol{\infty}}^{\boldsymbol{y}}\psi_i(\boldsymbol{z} - \boldsymbol{y}_i) \mathrm{d}\boldsymbol{z} \\
& = \sum\limits_{i=1}^{m} b_i\,\Psi_i(\boldsymbol{y} - \boldsymbol{y}_i). 
\end{split}
\end{equation}
The distribution functions $\Psi_i$ for the kernels $\psi_i$ are usually known or at least easy to calculate. The generation of the $k=1,\ldots,q$ random samples~(\ref{f5}) can be performed in three stages:
\begin{enumerate}
\item Choose randomly one of the $m$ kernels $\psi_i$ corresponding to the probabilities $b_i$. 
\item Let $j$ be the choice of the first stage. Generate now a $\psi_j$-distributed random value using the distribution function $\Psi_j$. 
\item Add the kernel center $\boldsymbol{y}_j$ to the random value from stage two to get a random sample $\boldsymbol{y}_k'$
\end{enumerate}

After that, we calculate the function values $h(\boldsymbol{y}_k')$ for all $k=1,\ldots,q$ of dataset~(\ref{f5}). The argument $\boldsymbol{y}_k'$ for which $h(\boldsymbol{y}_k')$ becomes maximal is then a good starting point for a local optimization method like gradient ascent \citep{Duda00}, for example.

\section{Implementation Example}

The subsequent Matlab code\footnote{The code was tested with Matlab 7.1.} snippet implements the described method for multidimensional Gaussian kernels with diagonal covariance matrix. 
\begin{verbatim}
function [xm,pxm] = findMax(para,q)

m = length(para.a);
d = length(para.x(1,:));

cdf = zeros(1,m);
for (i=2:m) cdf(i) = cdf(i-1) + para.a(i-1); end

xr = zeros(q,d); yr = zeros(q,1);
for (i=1:q)
    rv = rand(1); lvi = find(cdf < rv); ri = lvi(length(lvi));
    xr(i,:) = randn(1,d).*sqrt(para.s(ri)) + para.x(ri);
    yr(i) = KDE(xr(i,:),para);
end

[pxm,xi] = max(yr); xm = xr(xi,:);
\end{verbatim}
The parameters of the expression~(\ref{f6}) are combined into the structure \verb+para+ with three elements: \verb+para.a+ are the weights $b_i$, \verb+para.s+ the standard deviations, and \verb+para.x+ the centers of the Gaussian kernels. The function  \verb+KDE+ calculates the estimated density value for a given \verb+x+:
\begin{verbatim}
function y = KDE(x,para);

m = length(para.a); y = 0;
for (i=1:m)
    y = y + para.a(i)*gauss(x,para.x(i,:),para.s(i,:));
end

function y = gauss(x,m,s)

y = prod(1./(sqrt(2*pi)*s)).*exp(sum(-(x-m).^2./(2*s.^2)));
\end{verbatim}
The gradient ascent is not performed in this example. 

\begin{figure}[th]
\centerline{
\includegraphics[width=0.5\columnwidth]{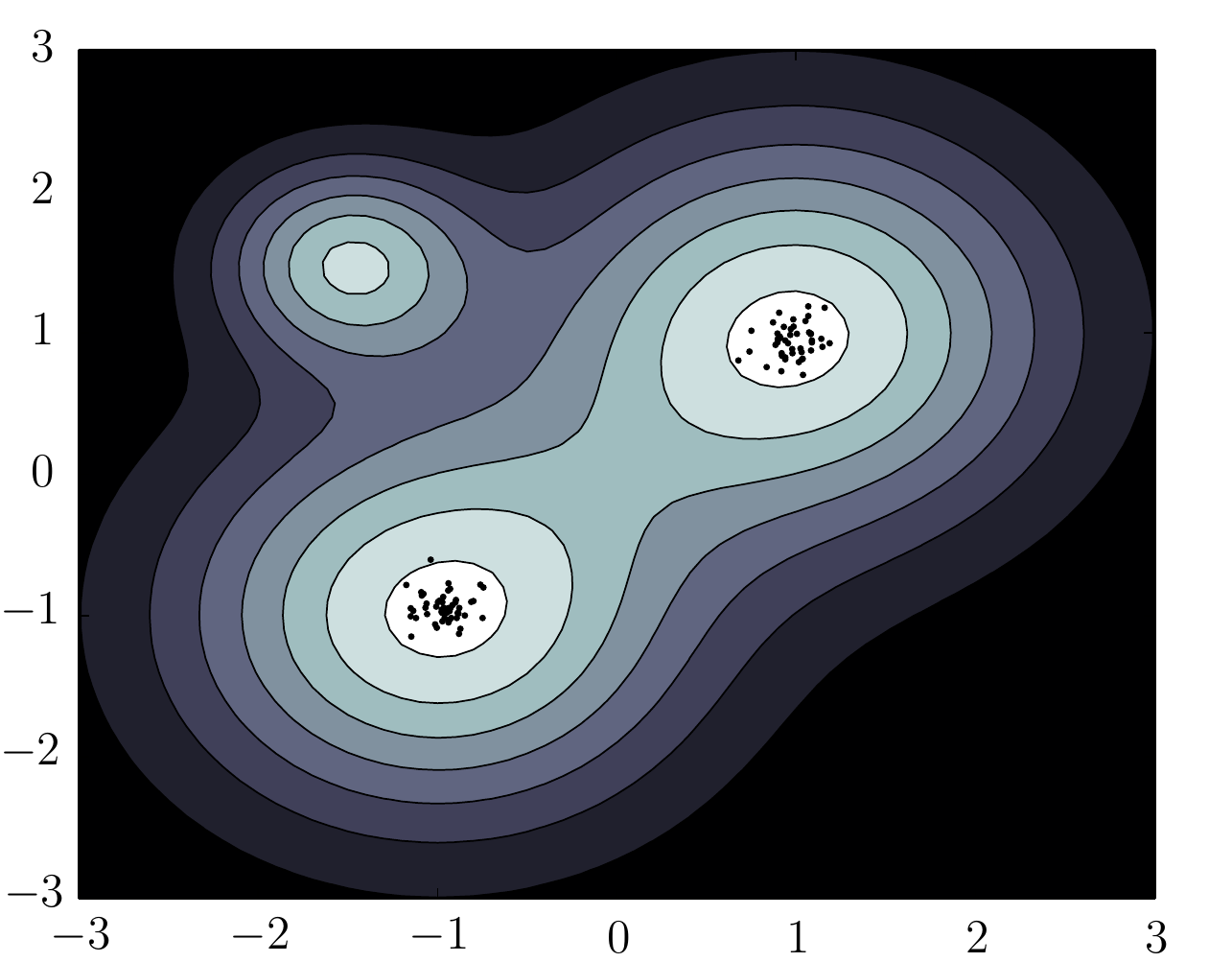}
\includegraphics[width=0.5\columnwidth]{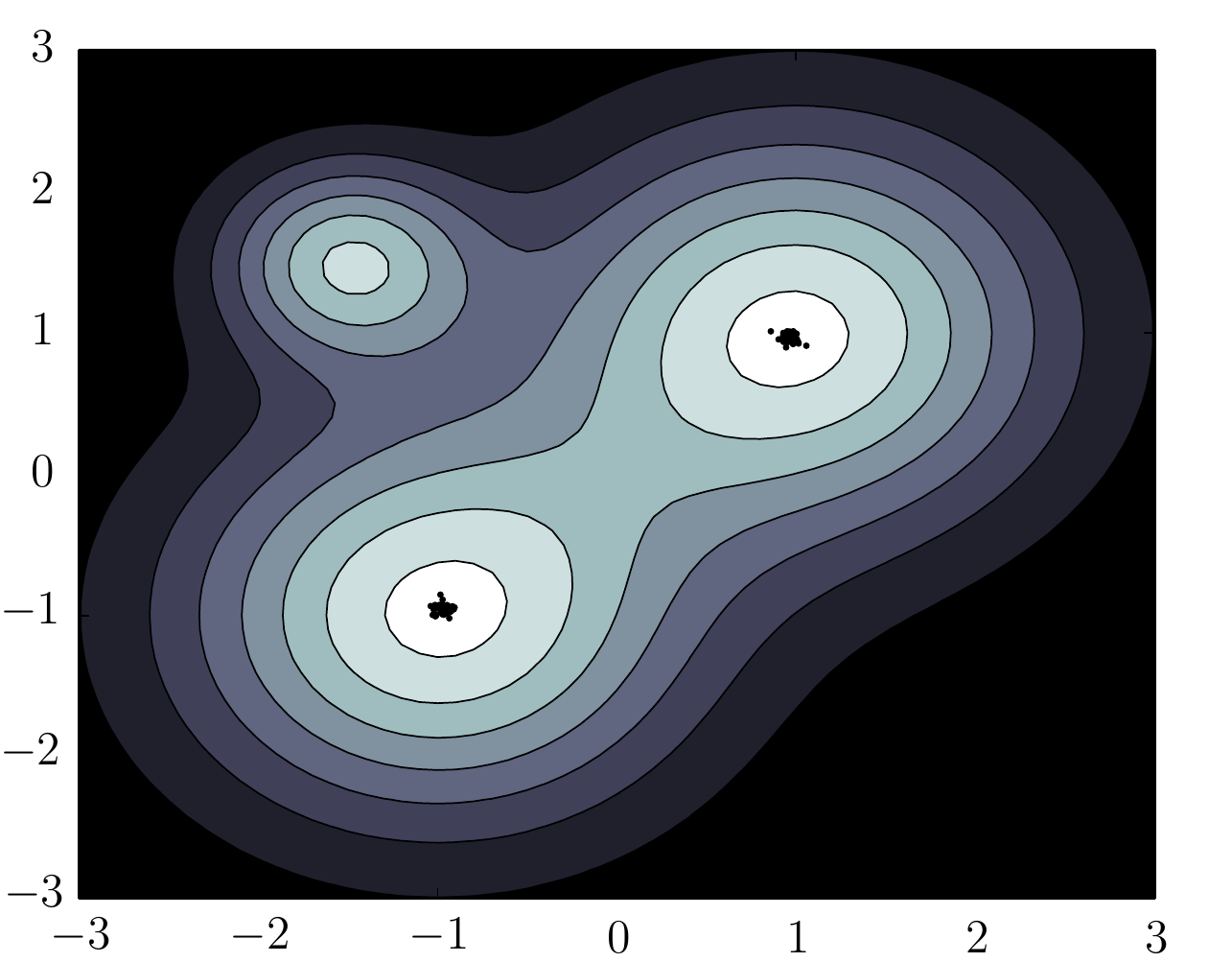}
}
\caption{Contour plots for a two-dimensional kernel density estimation with two global maxima at about $(1\,1)^T$ and $-(1\,1)^T$. Both were found by the algorithm (black marks). For the left hand plot, $q$ was $100$ and for the right hand plot $1000$.}
\label{fig2}
\end{figure} 

\begin{figure}[th]
\centerline{
\includegraphics[width=0.5\columnwidth]{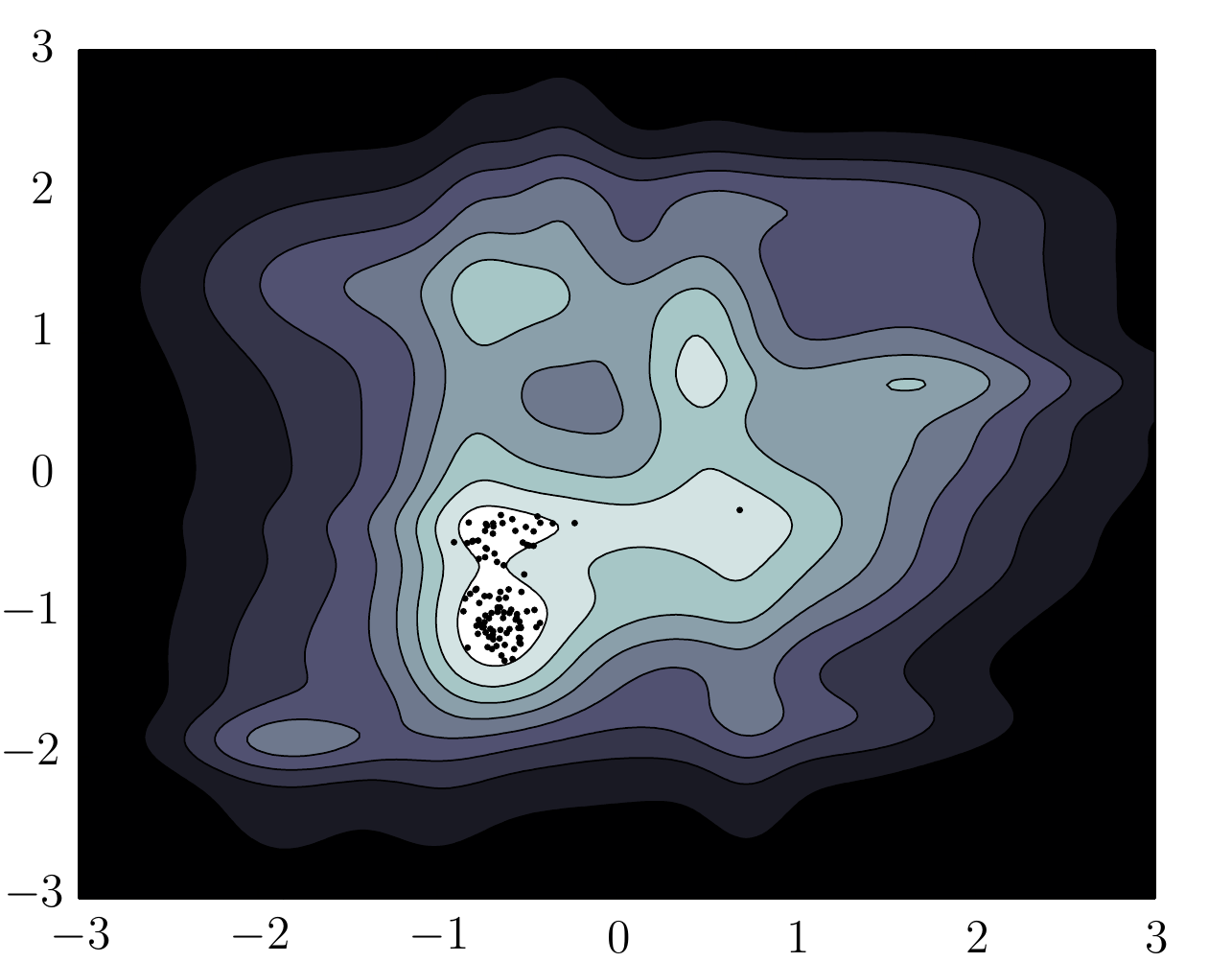}
\includegraphics[width=0.5\columnwidth]{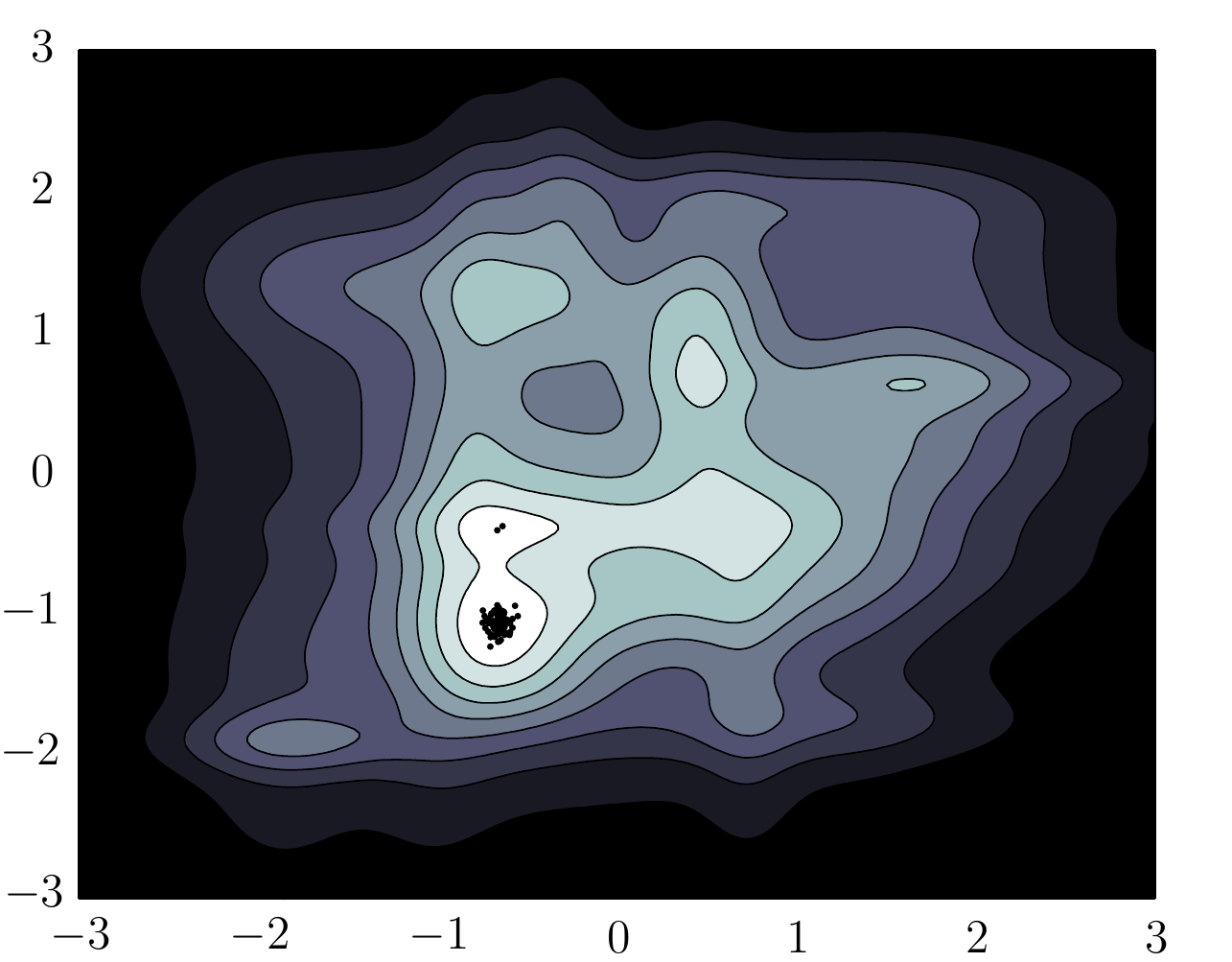}
}
\caption{The result for a more complex density with $80$ kernels and several local maxima. $q$ was $100$ (left) and $1000$ (right).}
\label{fig3}
\end{figure}

Figure~\ref{fig2} shows the results of an experiment with function \verb+findMax+ for different values of \verb+q+. The parameters of the probability density function $h(\boldsymbol{y})$ were
\begin{verbatim}
para.a = [0.45,0.45,0.1];
para.x = [[1,1];[-1,-1];[-1.5,1.5]];
para.s = [[1,1];[1,1];[0.5,0.5]];
\end{verbatim}
That means that there are two global maxima -- at approximately $(1\,1)^T$ and $-(1\,1)^T$. For this reason, both could be the result returned by the algorithm. But only one of these possibilities is returned per step. The plots also show that the distribution of the computed points becomes more compact with increasing size of $q$. Figure~\ref{fig3} shows the result for a more complex density with $80$ kernels and several local maxima.

\section{Regression Experiments}

\begin{figure}[th]
\centerline{
\includegraphics[width=0.5\columnwidth]{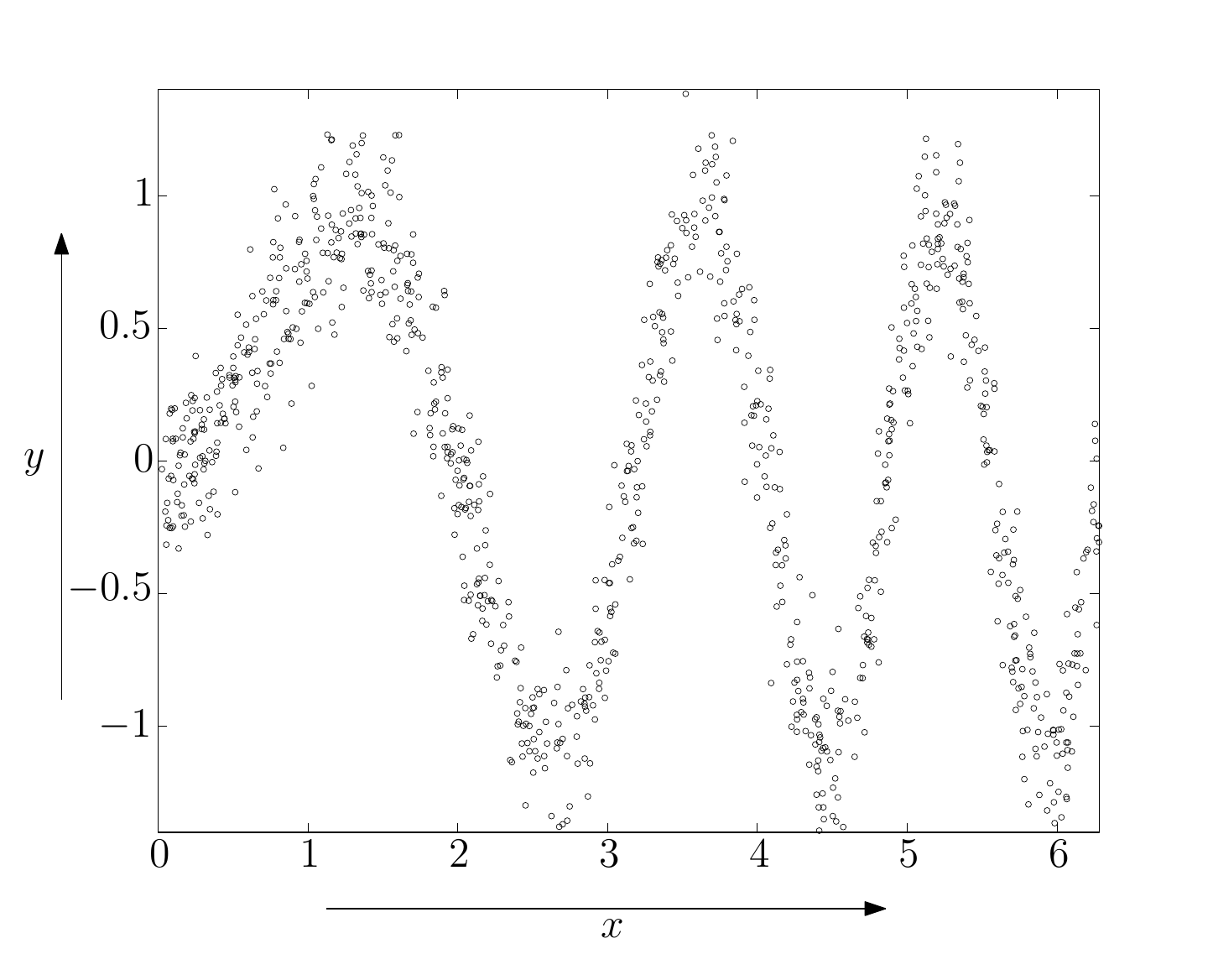}
\includegraphics[width=0.5\columnwidth]{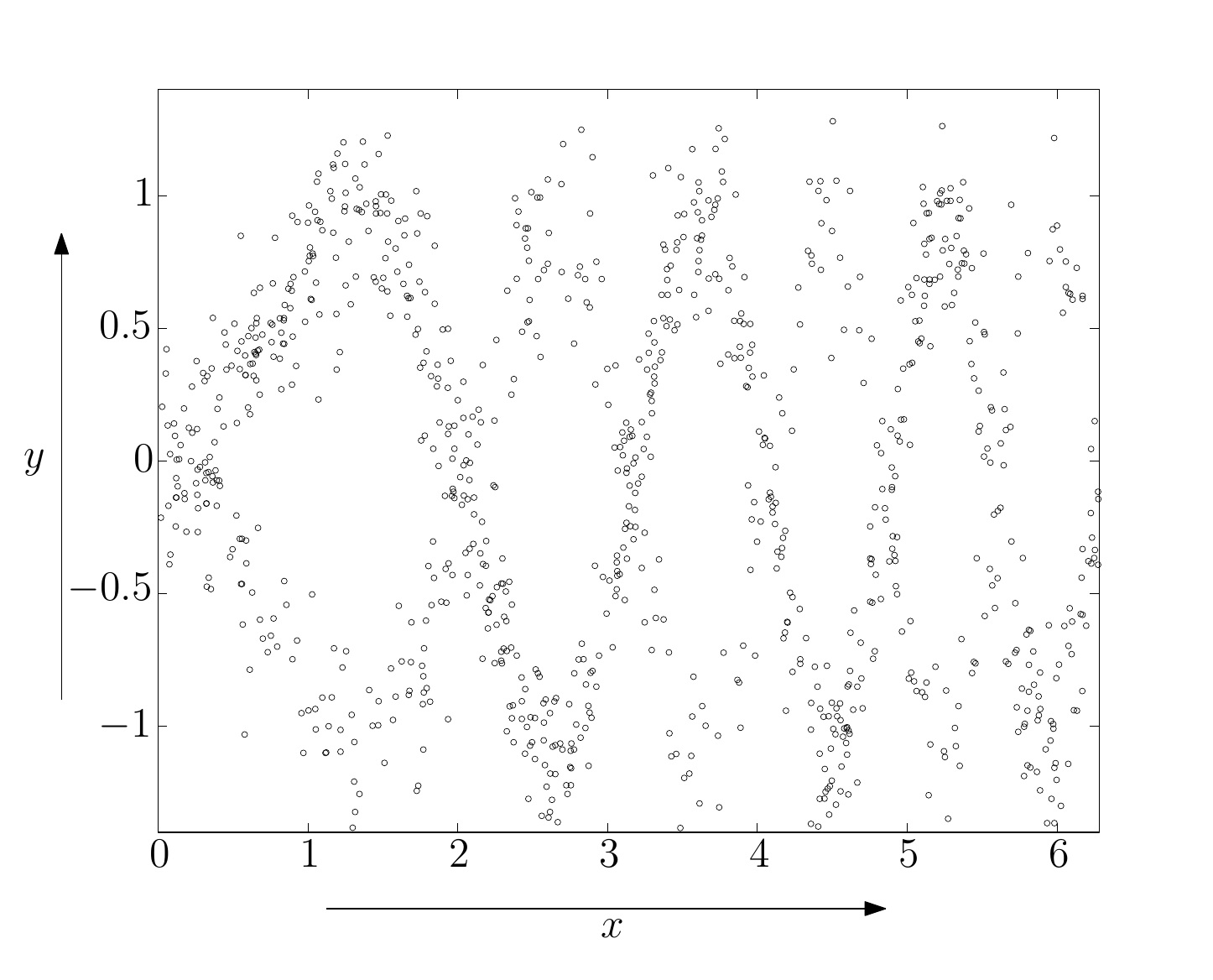}
}
\caption{Two datasets.}
\label{fig4}
\end{figure}

This section investigates the properties of the described method. The first experiment compares the standard Nadaraya-Watson method and the proposed method with computation of the mode in view of its ability to estimate a clear functional dependency between $\boldsymbol{x}$ und $\boldsymbol{y}$. For this purpose, a dataset of $n = 1000$ random sample points of the function $y = \sin(x^{\frac{8}{5}})$ in the interval $[0,2\,\pi]$ was generated. Furthermore, a slight, Gaussian distributed noise with a standard deviation of $\sigma_N = 0.2$ was added to the $y$-values. The dataset is shown on the left of Figure~\ref{fig4}.

Before applying the two regression methods, the distribution of the data has to be modeled by a kernel density. Different types of kernels can be applied. One of the simplest is the $d$-dimensional Gaussian kernel 
\begin{equation}
g(\boldsymbol{x},\boldsymbol{s}) = \prod\limits_{k=1}^{d}\frac{1}{\sqrt{2\,\pi} s_k} \exp\left(-\frac{x_k^2}{2\,s_k^2}\right)
\end{equation}
with $\boldsymbol{s} = (s_1\,\ldots\,s_d)^T$ as only free parameter. Its application to the two dimensional problem of Figure~\ref{fig4} yields
\begin{equation}
\tilde{p}(\boldsymbol{x}) = \frac{1}{n} \sum\limits_{i=1}^{n} g(\boldsymbol{x}-\boldsymbol{x}_i,\boldsymbol{s})
\end{equation}
with $\boldsymbol{x} = (x\,y)^T$.
For high-dimensional problems, the smoothness $\boldsymbol{s}$ has to be automatically optimized with respect to a certain quality measurement, such as the self-contribution \citep{Duin76} for example. Another method is plug-in estimation. A good overview about this topic is given by \citet{Turlach93} or \citet{Scott92}. 

\begin{figure}[th]
\centerline{
\includegraphics[width=0.5\columnwidth]{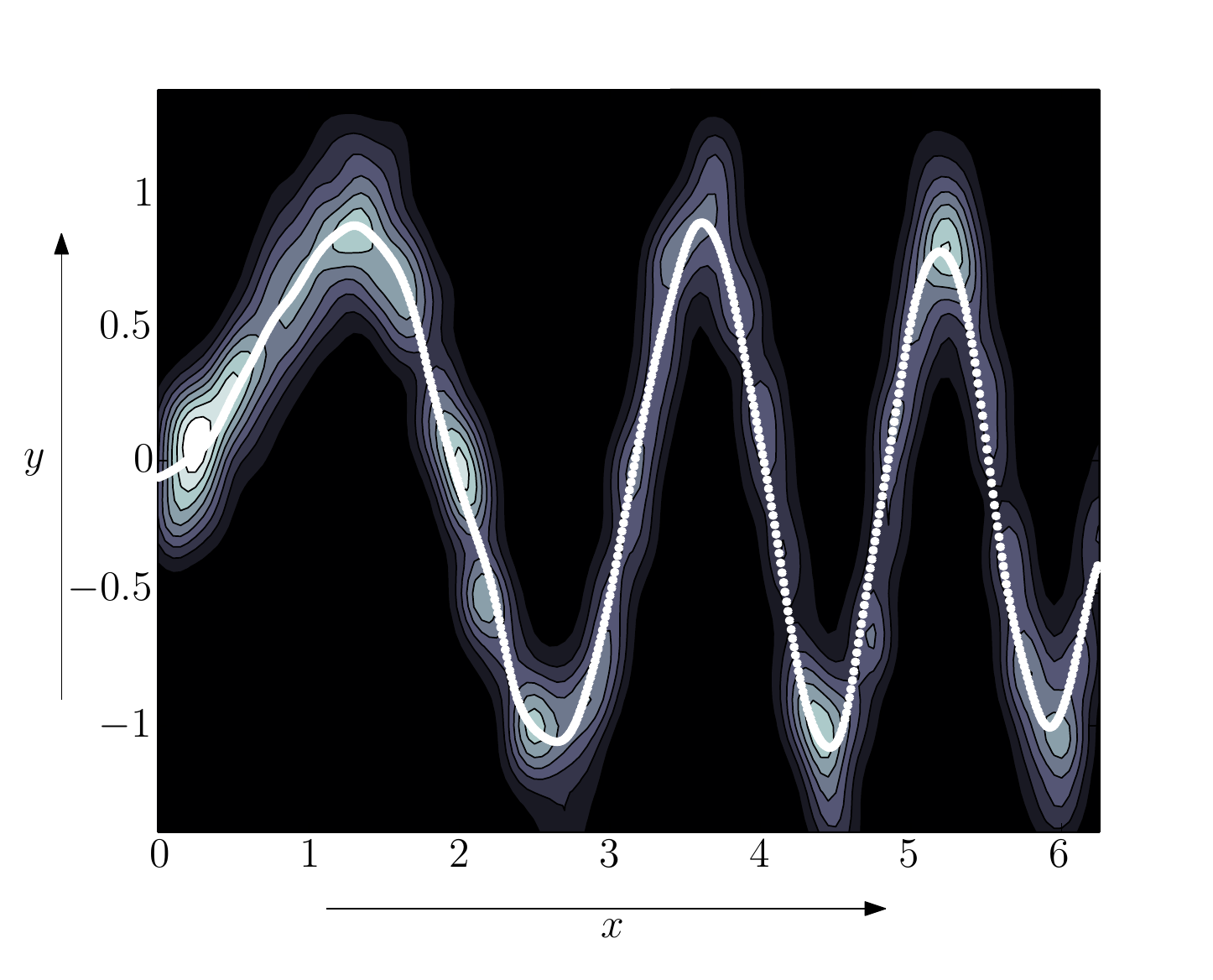}
\includegraphics[width=0.5\columnwidth]{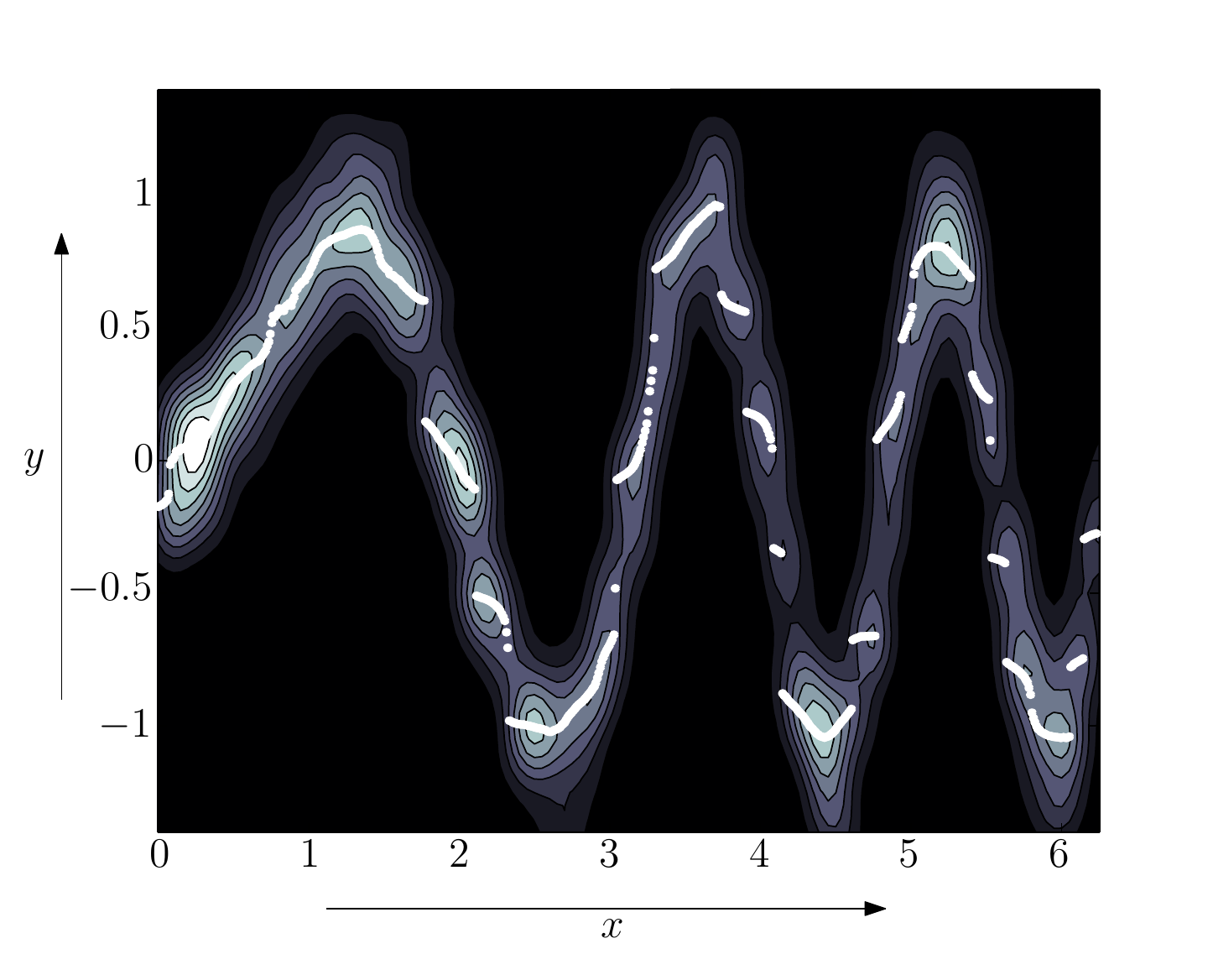}
}
\centerline{
\includegraphics[width=0.5\columnwidth]{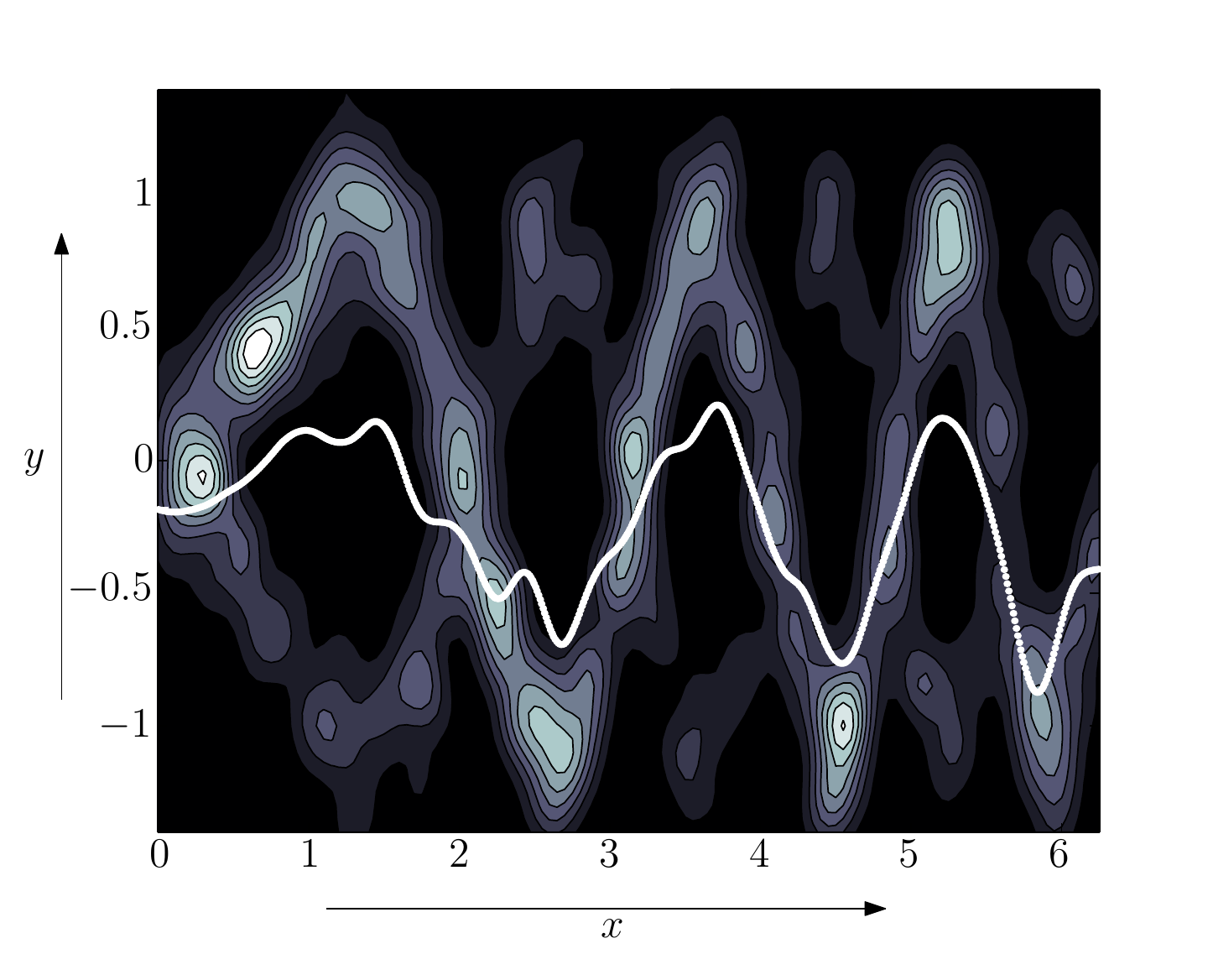}
\includegraphics[width=0.5\columnwidth]{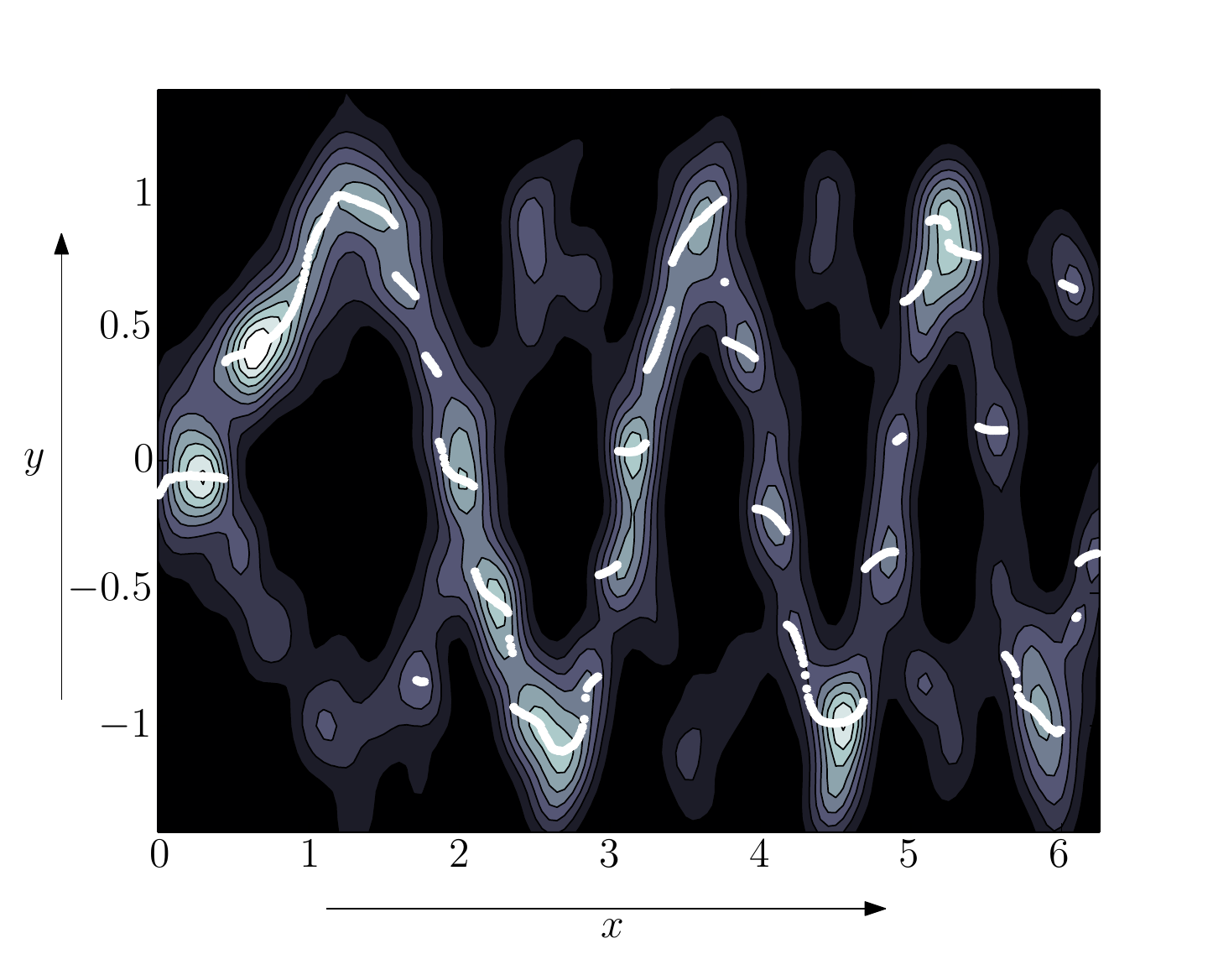}
}
\caption{The results of the Nadaraya-Watson method (left) and of the optimization method (right). The density estimation above shows a clear dependency between $x$ and $y$ -- contrary to the other below.}
\label{fig5}
\end{figure}

For the two-dimensional dataset in Figure~\ref{fig4}, it is still possible to estimate the smoothness parameter visually. For $\boldsymbol{s} = (0.1\,0.1)^T$, the resulting density is drawn as contour plot in Figure~\ref{fig5} at the top. Furthermore, the picture provides the result of the Nadaraya-Watson regression (left) and of the proposed method (right) as white dotted lines. Every point represents the outcome for a single given value $x$. 

The picture demonstrates too that the Nadaraya-Watson regression performs clearly better. This is only at the first glance surprising. Formally, both regression methods should give the same results, because expected value and maximum are identical for the true conditional probability distribution
\begin{equation}
p_{Y|X}(y|x) = g(y - \sin(x^{\frac{8}{5}}),\sigma_N).
\end{equation}
But both regression techniques have different susceptibilities to estimation errors and the property of the effective value to average out noise leads to a much smoother curve progression. 

This advantage becomes a shortcoming if the dependencies within the data are ambiguous. To demonstrate this, a second dataset was generated (Figure~\ref{fig4}, right). Now, for most values of $x$ two values of $y$ with different emphasis are reasonable. The Nadaraya-Watson regression calculates for every $x$ a ``compromise''. This can lead to very improbable values for $y$. The optimization method however chooses always the most probable value and is because of this immune to this effect. 

\section{Conclusion}

If a dependency between some data is clear and unambiguous, the standard Nadaraya-Watson method or -- still better -- the local linearizing Nadaraya-Watson approach \citep{Cleveland79} should be preferred for the modeling. But for numerous real life applications, it cannot be guaranteed that this condition is fulfilled because, for example, the data may be collected online. Another difficulty is a high dimensionality. The dependency may be simple and unambiguous between some of the vector components, but between others it may not. Every input-output combination has to be checked, what is mostly impracticable. 

For such general and complex cases, the proposed method is more suitable, because the assumption of unambiguousness is not necessary. The approach returns always a prediction that is probable in respect to the knowledge given by the sample data. For some applications, this property is more important than continuity of the curve and its smoothness. 

An example for such a scenario is machine control. The data are in this case measurements from actuators and sensors. The controller continuously has to solve the problem which actuator values leads to the desired sensor values. For this purpose, already one good setting is sufficient, regardless of whether several possibilities exist or not. An average value or a ``compromise'', however, is mostly a bad decision. 

\addcontentsline{toc}{section}{References}
\bibliographystyle{plainnat}

\end{document}